\documentclass[conference]{IEEEtran}
\IEEEoverridecommandlockouts

\usepackage{cite}
\usepackage{amsmath,amssymb,amsfonts}
\usepackage{algorithmic}
\usepackage{graphicx}
\usepackage{textcomp}
\usepackage{xcolor}
\usepackage{soul}
\usepackage{comment,url,enumitem}
\usepackage{breqn}
\usepackage{tablefootnote}

\usepackage{multirow,booktabs,pifont,amsmath,array}

\usepackage[utf8]{inputenc}
\usepackage[T1]{fontenc}

\def\BibTeX{{\rm B\kern-.05em{\sc i\kern-.025em b}\kern-.08em
    T\kern-.1667em\lower.7ex\hbox{E}\kern-.125emX}}
\begin{document}

\title{Serialized Output Prompting for Large Language Model-based Multi-Talker Speech Recognition
}

\author{
\IEEEauthorblockN{
\textit{Hao Shi, Yusuke Fujita, Tomoya Mizumoto, Lianbo Liu, Atsushi Kojima, Yui Sudo}
}
\vspace{0.1cm}
\IEEEauthorblockA{
{SB Intuitions, Tokyo, Japan}
}
}

\maketitle

\begin{abstract}
Prompts are crucial for task definition and for improving the performance of large language models (LLM)-based systems. 
However, existing LLM-based multi-talker (MT) automatic speech recognition (ASR) systems either omit prompts or rely on simple task-definition prompts, with no prior work exploring the design of prompts to enhance performance. 
In this paper, we propose extracting serialized output prompts (SOP) and explicitly guiding the LLM using structured prompts to improve system performance (SOP-MT-ASR). 
A Separator and serialized Connectionist Temporal Classification (CTC) layers are inserted after the speech encoder to separate and extract MT content from the mixed speech encoding in a first-speaking-first-out manner. 
Subsequently, the SOP, which serves as a prompt for LLMs, is obtained by decoding the serialized CTC outputs using greedy search. 
To train the model effectively, we design a three-stage training strategy, consisting of serialized output training (SOT) fine-tuning, serialized speech information extraction, and SOP-based adaptation. 
Experimental results on the LibriMix dataset show that, although the LLM-based SOT model performs well in the two-talker scenario, it fails to fully leverage LLMs under more complex conditions, such as the three-talker scenario. 
The proposed SOP approach significantly improved performance under both two- and three-talker conditions. 
\end{abstract}

\begin{IEEEkeywords}
automatic speech recognition, large language model, multi-talker, prompt
\end{IEEEkeywords}

\section{Introduction}
\label{section:intro}
Multi-talker (MT) automatic speech recognition (ASR) \cite{shi2024advancing,shi_slt2024,meng2024large,8682822,8461893} aims to transcribe all talkers' speaking contents from overlapping speech. 
It's inherently more challenging than standard ASR \cite{10038197,10542371,9689650,yang2024large,shu24_interspeech,song22e_interspeech,zhao2024adapting} due to the difficulty of recognizing MT overlapping speech. 
MT-ASR systems have evolved from the separation-first-then-recognition pipeline \cite{6739096,10448116,8369155,10542371} to approaches that rely solely on the end-to-end ASR back-end to handle MT speech overlap \cite{7979557,9054328}. 
Utterance-level permutation invariant training (uPIT) \cite{7979557} is a widely used training strategy for MT-ASR \cite{7952154}. 
The smallest loss among all possible permutations of multiple outputs is used for backpropagation \cite{7952154}. 
The computational complexity increases significantly as the number of speakers grows \cite{kanda20b_interspeech}. 
Serialized output training (SOT) \cite{kanda20b_interspeech} is proposed to address the aforementioned limitations. 
SOT organizes training labels by serializing overlapping speech into a single token sequence based on the speaking start time of each talker \cite{kanda20b_interspeech}. 
It alleviates the issue of variable talker numbers without performance degradation compared to uPIT-based ASR \cite{kanda20b_interspeech}.

Recently, SOT MT-ASR models have been integrated with large language models (LLM) \cite{liu2019roberta,kenton2019bert,radford2018improving,lewis2019bart,yang2019xlnet,hu2024wavllm,touvron2023llama,brown2020language} , demonstrating impressive performance improvement \cite{shi2024advancing,meng2024large}. 
SOT-based ASR is based on the attention-based encoder-decoder (AED) structure \cite{6af3452a28a04980b2b8f5eb48730d36,chan2015listen}. 
Thus, LLMs can be easily incorporated into SOT-based ASR as the decoder. 
Powerful LLM-based decoders help improve poor grammatical structures in sentences and necessitate strong long-context awareness and cross-utterance modeling \cite{shi2024advancing}.

However, LLM-based SOT models still struggle in complex overlapping scenarios, such as simultaneous speech by three speakers. 
LLMs are not pretrained with scenes where multiple text contents overlap, making it difficult to handle such scenarios without adaptation. 
Besides, supervised finetuning on limited MT training data offers a partial solution, but it does not fully exploit the LLM's adaptability. 
While prompting is known as a critical component to fully utilize the LLM's adaptability, we found that existing LLM-based SOT models are used with a static prompt that merely specifies the MT-ASR task itself \cite{meng2024large}.

In this paper, we propose an LLM-based MT-ASR with prompting for improving performance in complex overlapping scenarios. 
We use an adaptive prompt to indicate how MT contents are mixed and can be separated according to the input. 
To generate the prompt, we introduce serialized Connectionist Temporal Classification (CTC)-based ASR as an auxiliary network.  
An additional Separator \cite{shi_slt2024} and speaking-time-aligned CTC layers are inserted after the speech encoder to extract MT speaking content from the mixed speech encoding. 
The number of CTC layers is equal to the number of talkers, and the sequence of CTC outputs is ordered according to the talkers' speaking start times. 
The greedy search results of the serialized CTC are referred to as serialized output prompting (SOP), which are used as prompts for LLM decoding.

To train the model effectively, we propose a three-stage training strategy. 
In the first stage, the model is trained with SOT to enable the speech encoder to encode the mixture inputs and to adapt the LLM decoder to the mixed speech representations. 
Then, in the second stage, the speech encoder, along with the Separator and serialized CTC layers, is trained to extract the SOP. 
This is to prevent potential non-differentiability issues caused by the CTC layers during joint training with the decoder in the utilization of SOP. 
However, training the serialized CTC layers and speech encoder degrades the LLM decoding performance. 
Therefore, another group of adapters \cite{hu2022lora} is introduced into the LLM for SOP adaptation. 
The speech encoder, separator, and CTC layers are frozen during the third stage.

The remainder of this paper is organized as follows. 
Section~\ref{sec:prelim} introduces the preliminaries. 
The proposed method is detailed in Section~\ref{sec:proposed}. 
Experimental settings and results are reported in Section~\ref{sec:exp}. 
Finally, Section~\ref{sec:conclu} concludes the paper.

\section{Preliminaries}
\label{sec:prelim}

\subsection{LLMs as Decoder for ASR}
The structure for LLM-based ASR comprises a speech encoder, downsampling layers, a projector, and the LLM-based decoder. 
Pretrained models \cite{9814838, 10096630, radford2023robust} are often used for downstream tasks. 
The speech signal $\textbf{y}$ is first converted into a speech encoding: 
\begin{equation}
    \textbf{H}_e = \text{Enc}(\textbf{y}),
\end{equation}
where $\textbf{H}_e$ represents the speech encoding. 
$\text{Enc}$ represents the speech encoder. 
$\textbf{H}_e$ is typically several times longer than the final text transcription, which increases computational complexity and demands greater processing capability from LLMs. 
Thus, some downsampling strategies are adopted. 
Two common methods are used for downsampling: the first involves several 2D convolutional layers, while the second concatenates $n$ consecutive frames along the feature dimension: 
\begin{equation}
    \textbf{H}_d = \text{Down}(\textbf{H}_e),
\end{equation}
where $\textbf{H}_d$ represents the downsampled speech encoding. 
$\text{Down}$ represents the downsampling layer. 
After downsampling, the projector performs the dimension conversion between speech encoding and text representation, which can be represented as follows: 
\begin{equation}
    \textbf{H}_p = \text{Projector}(\textbf{H}_d),
\end{equation}
where $\textbf{H}_p$ represents the projected encoding. 
Linear layers are commonly used as the projector.

Finally, the LLM-based decoder transcribes the text according to the projected encoding: 
\begin{equation}
    \textbf{T}_e = \text{LLM}([\textbf{P}], \textbf{H}_p, \textbf{E}_t).
    \label{conventinal_llm_input}
\end{equation}
where $\textbf{E}_t$ represents the text embedding. 
$[\star]$ indicates that the component $\star$ is optional and may or may not be used. 
Conventional LLM-based MT-ASR systems use either no prompt or a simple task prompt $\textbf{P}$, which is concatenated before $\textbf{E}_t$. 
$\textbf{T}_e$ represents the decoding transcription. 
The projected encoding is concatenated with the text embedding to serve as the input to the decoder. 
During finetuning, the LLMs are typically frozen, with only the inserted adapters \cite{hu2022lora,shi2024investigation,shi24b_interspeech} being trainable. 
Cross-Entropy (CE) is used as the loss function:
\begin{equation}
    \mathcal{L} = \text{CE}(\textbf{T}_{l}, \textbf{T}_{e}).
\end{equation}
where $\textbf{T}_{l}$ represents the label.

\subsection{Serialized Output Training (SOT) for MT-ASR}
SOT arranges the transcriptions of multiple talkers sequentially based on their speaking start times to create a unified transcription.
A special symbol, $\langle sc \rangle$, is inserted between the transcriptions of different talkers to indicate speaker change. 
For instance, in the case of two talkers, the target sequence is represented as $\textbf{T}_{sot} = \{t_1^{1}, \dots, t_1^{N^{1}}, \langle sc \rangle, t_2^{1}, \dots, t_2^{N^{2}} \}$, where $t_1$ and $t_2$ denote the transcriptions of the first-speaking and second-speaking talkers, respectively. 
The $N^{1}$ and $N^{2}$ represent their transcriptions lengths.

With this training target, the attention mechanism can effectively focus on the relevant portions of overlapping speech encoding and decode the transcriptions $\textbf{T}_e$ of multiple talkers sequentially according to their speaking times. 
The loss function for SOT-based ASR can be represented as follows: 
\begin{equation}
    \mathcal{L}_{\text{SOT}} = \text{CE}(\textbf{T}_e, \textbf{T}_{sot}),
    \label{eq:sot_ce}
\end{equation}
Only CE loss is used during the training of the SOT-based ASR system.

\begin{figure*}
    \centering
    \includegraphics[width=0.87\linewidth]{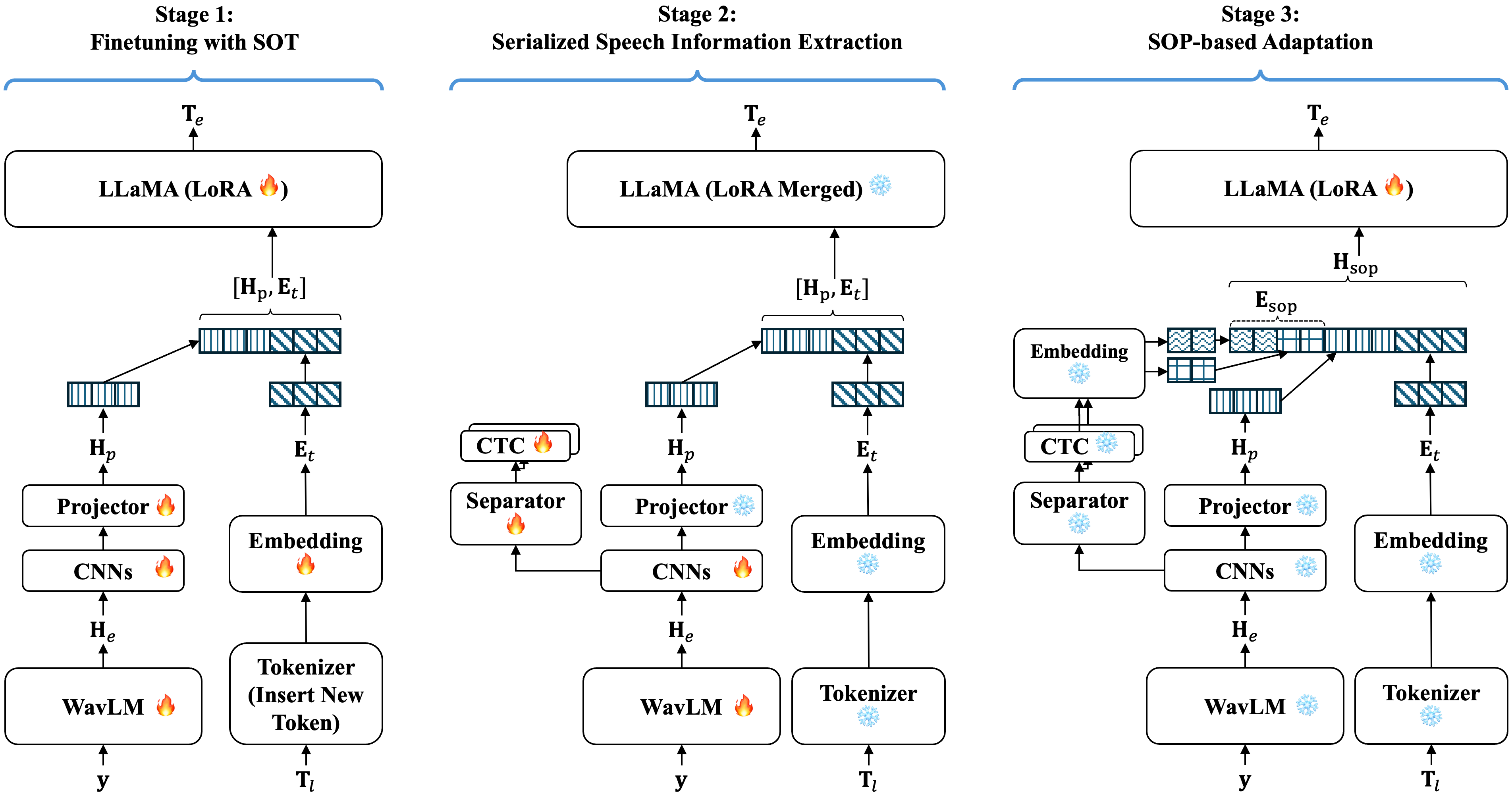}
    \vspace{-10pt}
    \caption{
    The flowchart of the proposed SOP for LLM-based MT-ASR. 
    It contains three training stages: (1) fine-tuning with SOT, (2) serialized speech information extraction, and (3) SOP-based adaptation. 
    }
    \label{fig:enter-label}
    \vspace{-10pt}
\end{figure*}

\subsection{Single-Talker Information Guidance SOT}
\label{sec:gencsep}
To improve the encoder's representation, the overlapped encoding separation (EncSep) \cite{shi_slt2024} is proposed to utilize the Connectionist Temporal Classification (CTC)-Attention hybrid loss in the SOT-based ASR. 
A Separator is introduced to disentangle the mixed embedding $\textbf{H}_e$ into individual talker-specific representations $\textbf{H}_{sep}^{1}, ..., \textbf{H}_{sep}^{S}$, $S$ represents the number of talkers: 
\begin{equation}
    \textbf{H}_{sep}^{1}, ..., \textbf{H}_{sep}^{S} = \text{Separator}(\textbf{H}_{e}).
\end{equation}
The Long Short-Term Memory (LSTM) \cite{graves2012long} is adopted as the separator: 
\begin{equation}
    \begin{split}
        \textbf{H}_{sep}^{s} = \text{ReLU}(\text{Linear}^{s}(\text{LayerNorm}(\text{LSTM}(\textbf{H}_{e})))),
    \end{split}
\end{equation}
Multiple linear layers are employed to extract the single-talker representations $\textbf{H}_{sep}^{s}$. 
Each linear layer is associated with a specific talker, determined by the serialized order based on their speaking onset times. 
The computation of the serialized CTC loss is then performed as follows: 
\begin{equation}
    \mathcal{L}_{\text{CTC-EncSep}} = \sum\nolimits_{s=1}^{S}\text{Loss}_{\text{CTC}}(\mathbf{H}_{sep}^{s}, \mathbf{T}^{s})
\end{equation}
where $\textbf{T}^{s}$ denotes the transcription corresponding to the $s$-th speaker, ordered according to the serialized speaking sequence. 
In addition, the CE loss, as defined in Eqn.~(\ref{eq:sot_ce}), is incorporated into the training objective. 
The overall training objective for EncSep is defined as follows: 
\begin{equation}
    \mathcal{L}_{\text{EncSep}} = \alpha \mathcal{L}_{\text{CTC-EncSep}} + (1 - \alpha)\mathcal{L}_{\text{SOT}},
    \label{eq:encsep}
\end{equation}
where $\alpha$ is a tunable hyperparameter controlling the trade-off between the two loss components.

The single-talker information guidance SOT (GEncSep) further utilizes the separated embeddings $\textbf{H}_{sep}^{1}, ..., \textbf{H}_{sep}^{S}$. 
The separator decomposes the overlapped embedding $\textbf{H}_{e}$ into individual talker-specific embeddings $\textbf{H}_{sep}^{1}, \ldots, \textbf{H}_{sep}^{S}$. 
The separated embeddings are subsequently concatenated along the time dimension as follows: 
\begin{equation}
    \textbf{H}_{con} = \text{Concat}(\textbf{H}_{sep}^{1}, ..., \textbf{H}_{sep}^{S})
\end{equation}
The attention mechanism is employed to calculate attention weights conditioned on the single-talker information as follows: 
\begin{equation}
    \textbf{a}_{con}^{n} = \text{Attention} \left( \textbf{H}_{con}, \textbf{d}^{n-1} \right).
\end{equation}
$\textbf{a}_{con}^{n}$ denotes the context vector derived from the concatenated embedding $\textbf{H}_{con}$ using the attention mechanism. 
$\textbf{d}^{n-1}$ is the hidden state of the decoder. 
The decoder generates predictions based on both the attention-derived features and the previously generated tokens: 
\begin{equation}
    \textbf{c}^{n} = \text{Decoder} \left( \textbf{H}_{con}, \textbf{a}_{con}^{n}, \textbf{c}^{1:n-1} \right). 
\end{equation}
The decoder generates the corresponding output sequence $\textbf{C}$ in an iterative manner. 
The training objective for GEncSep follows the same formulation as specified in Eqn.~(\ref{eq:encsep}).

\section{Proposed Method: Serialized Output Prompting for LLM-based MT-ASR}
\label{sec:proposed}
The performance of LLMs is highly influenced by prompt design, which plays a crucial role in shaping model behavior and guiding contextual interpretation. 
However, the existing LLM-based MT-ASR systems use either no prompts or only simple task-definition prompts (shown in Eqn.~(\ref{conventinal_llm_input})). 
In this work, we propose serialized output prompting (SOP), a method that explicitly guides LLM-based MT-ASR through structured prompts. 
To extract talker-specific content from overlapped speech, we introduce two additional components: a Separator \cite{shi_slt2024} and serialized Connectionist Temporal Classification (CTC) layers aligned with talker start times. 
These modules are inserted after the speech encoder to generate serialized contents. 
Each CTC layer corresponds to one speaker, and the outputs are temporally ordered based on when each talker begins speaking. 
The resulting token sequences, obtained via greedy decoding from the CTC layers, form the SOP, which are then provided as prompts to the LLM decoder to improve ASR accuracy under multi-talker conditions. 
The overall architecture is shown in Fig.~\ref{fig:enter-label}--(Stage~3).

\subsection{Overall of SOP for LLM-based MT-ASR}
The proposed method consists of a speech encoder, several downsampling layers, a projector, a separator, multiple CTC layers, and an LLM-based decoder. 
The input speech signal $\textbf{y}$ is first used to extract frame-level acoustic features as follows: 
\begin{equation}
    \textbf{H}_e = \text{Enc}(\textbf{y}),
\end{equation}
where $\textbf{H}_e$ represents the extracted speech embedding. 
The extracted speech embeddings are passed through downsampling layers. 
Here, three Convolutional Neural Networks (CNNs) are used for downsampling: 
\begin{equation}
    \begin{aligned}
        & \textbf{H}^{(2)} = \text{Conv}_2(\text{Conv}_1(\textbf{H}_e)), \\
        & \textbf{H}^{(3)} = \text{Conv}_3(\textbf{H}^{(2)}), \\
        & \textbf{H}_{\text{d}} = \textbf{H}^{(3)},
    \end{aligned}
\end{equation}
where $\textbf{H}^{(2)}$ represents the embedding processed by the first and second CNN layers. 
$\textbf{H}^{(3)}$ represents the embedding processed by the third CNN layer, which also serves as the output of the down-sampling layers. 
Each of the CNN layers performs a two-times downsampling.

An LSTM-based Separator disentangles overlapping speech into talker-specific embeddings: 
\begin{equation}
    \textbf{H}_{{sep}} = \text{Separator}(\textbf{H}^{(2)}),
\end{equation}
Multiple linear layers are employed to extract the single-talker representations $\textbf{H}_{sep}^{s}$: 
\begin{equation}
    \begin{split}
        \textbf{H}_{sep}^{s} = \text{ReLU}(\text{Linear}^{s}(\text{LayerNorm}(\textbf{H}_{{sep}}))),
    \end{split}
\end{equation}
It should be noted that the output of the second CNN layer is used as the input to the Separator, instead of the final layer, since the final layer applies excessive downsampling. 
Although the output of the speech encoder $\textbf{H}_e$ could be used for the Separator, the sequence is relatively long, which increases computation time and CTC merging time. 
As a result, the output of the second CNN layer is chosen to extract talker information. 

This greedy decoding strategy provides an efficient means of converting the frame-level token predictions into a valid transcription: 
\begin{equation}
    \bar{\textbf{C}}^s = \text{Greedy}(\textbf{H}_{sep}^{s}),
\end{equation}
where $\bar{\textbf{C}}^s$ denotes the output sequence decoded from the $s$-th CTC branch. 
The serialized CTC output sequences are concatenated as the SOP: 
\begin{equation}
    \bar{\textbf{C}}_{sop} = \text{Concat}(\bar{\textbf{C}}^1, \bar{\textbf{C}}^2, ..., \bar{\textbf{C}}^s). 
\end{equation}
Then, the SOP is subsequently converted into embedding sequences as follows: 
\begin{equation}
    \textbf{E}_{\text{sop}} = \text{Embedding}(\bar{\textbf{C}}_{sop}). 
\end{equation}
where $\text{Embedding}$ represents the embedding layer. 
$\textbf{E}_{\text{sop}}$ represents the embedding sequence extracted from $\bar{\textbf{C}}_{sop}$.

The projector is used to align the mixture of speech encoding and text modalities, and to match their dimensional representations: 
\begin{equation}
    \textbf{H}_p = \text{Projector}(\textbf{H}_{\text{d}}).
\end{equation}
The SOP representations $\textbf{E}_{\text{sop}}$, the mixture speech embedding $\textbf{H}_{p}$, and the text embedding $\textbf{E}_t$ are concatenated to form the decoder input as follows, and shown  in Fig.~\ref{fig:enter-label}--(Stage~3): 
\begin{equation}
    \textbf{H}_{\text{sop}} = [\textbf{E}_{\text{sop}}; \textbf{H}_p ; \textbf{E}_t],
\end{equation}
where $[\cdot ; \cdot]$ denotes the concatenation operation. 
$\textbf{E}_t$ represents the text embedding. 
The main difference between $\bar{\textbf{C}}_{sop}$ and $\textbf{P}$ in Eqn.~(\ref{conventinal_llm_input}) is that $\bar{\textbf{C}}_{sop}$ preserves the alignment with multi-speaker overlapping speech, rather than merely defining the task. 
The concatenated features $\textbf{H}_{\text{con}}$ are then fed into a LoRA-adapted LLM decoder to generate the serialized output sequence $\textbf{T}_e$: 
\begin{equation}
    \textbf{T}_e = \text{LLM}(\textbf{H}_{\text{sop}}, \textbf{LoRA}).
\end{equation}
where $\textbf{LoRA}$ represents the LoRA-based adapter \cite{hu2022lora}.

\subsection{Multi-Stage Training Strategies}
\label{subsec:ms-training}
We attempted to train all parameters directly in a single stage, but the system performance was suboptimal. 
During the experiments, we found that CTC training affects the performance of the LLM-based decoder when they are trained simultaneously. 
Besides, the Separator, CTC layers, and speech encoder can be directly trained effectively in the two-talker condition; however, in the three-talker condition, the Separator and CTC layers need to be trained using a speech encoder that has been finetuned with SOT. 
To achieve effective and stable model optimization, we adopt a multi-stage training strategy in which the model is gradually exposed to increasing levels of task complexity.

\subsubsection{Stage 1, finetuning with SOT}
This stage focuses on training the speech encoder to encode mixed speech embeddings and enabling the LLM-based decoder to learn the ability to serialize the mixed token sequences. 
The training loss remains the same as defined in Eqn.~(\ref{eq:sot_ce}). 
After training, the LoRA weights are merged into the LLM.

\subsubsection{Stage 2, serialized speech information extraction}
The Separator and CTC modules are introduced into the architecture to enable the model to handle overlapped speech scenarios. 
This stage extends the learning objective to include speaker-aware feature disentanglement and alignment supervision. 
The training now jointly optimizes: the speech encoder, CNN-based downsampling layers, the newly inserted Separator, and serialized CTC layers. 
The training loss is defined as in Eqn.~(\ref{eq:encsep}), which is applied not only to the CTC branch, but also to the LLM output.

\subsubsection{Stage 3, SOP-based adaptation}
In the final stage, we introduce an additional set of LoRA modules specifically designed to adapt the LLM to the SOP-based prompts produced by the Separator. 
Only the LoRA parameters are trainable. 
The training loss remains the same as defined in Eqn.~(\ref{eq:sot_ce}).

\section{Experiments}
\label{sec:exp}
\subsection{Datasets}
We used the LibriMix dataset \cite{cosentino2020librimix} to evaluate the model performance. 
It used the train-clean-100, train-clean-360, dev-clean, and test-clean subsets from the LibriSpeech dataset \cite{7178964} as the clean speech. 
For the noisy LibriMix, the noise samples were taken from WHAM! dataset \cite{Wichern2019WHAM}. 
We used the official scripts\footnote{\url{https://github.com/JorisCos/LibriMix}} to synthesize Libri2Mix and Libri3Mix. 
We used the offset file to make different speaking start times for multiple speakers. 
The two-speaker offset files follow the official ESPnet setting\footnote{\url{https://github.com/espnet/espnet/tree/master/egs2/librimix/sot_asr1}}, while the three-speaker offset files were created by ourselves and will be released after the anonymous review phase. 
Libri2Mix training set contains approximately 270 hours of speech, both the validation and evaluation sets contain about 11 hours each. 
Libri3Mix training set contains approximately 186 hours of speech, both the validation and evaluation sets also containing about 11 hours each.

\subsection{Model Configurations}
All the experiments were conducted using the Hugging Face packages. 
For the speech encoder, WavLM-Large\footnote{\url{https://huggingface.co/microsoft/wavlm-large}} was used, as WavLM includes MT data in its pre-training. 
For the LLM-based decoder, different sizes of LLaMA were used: LLaMA-3.2-1B\footnote{\url{https://huggingface.co/meta-llama/Llama-3.2-1B}}, LLaMA-3.2-3B\footnote{\url{https://huggingface.co/meta-llama/Llama-3.2-3B}}, and LLaMA-3.1-8B\footnote{\url{https://huggingface.co/meta-llama/Llama-3.1-8B}}.

\begin{table*}[]
\renewcommand{\arraystretch}{1.1}
\caption{The performance of the proposed method on the LibriMix datasets: Word Error Rate (WER) is used for evaluation.}
\centering
\begin{tabular}{c|c|c|l|l|cc|cc|cc|cc}
\toprule[1.5pt]
\multirow{3}{*}{\textbf{ID}} & \multirow{3}{*}{\textbf{\begin{tabular}[c]{@{}c@{}}\#Param. \\ (LLM)\end{tabular}}} & \multirow{3}{*}{\textbf{Stage}} & \multirow{3}{*}{\textbf{Systems}} & \multirow{3}{*}{\textbf{Input of LLMs}} & \multicolumn{4}{c|}{\textbf{Noisy}} & \multicolumn{4}{c}{\textbf{Clean}} \\
\cline{6-13}
& & & & & \multicolumn{2}{c|}{\textbf{Libri2Mix}} & \multicolumn{2}{c|}{\textbf{Libri3Mix}} & \multicolumn{2}{c|}{\textbf{Libri2Mix}} & \multicolumn{2}{c}{\textbf{Libri3Mix}} \\
\cline{6-13}
& & & & & \textbf{Dev} & \textbf{Eval} & \textbf{Dev}      & \textbf{Eval} & \textbf{Dev} & \textbf{Eval} & \textbf{Dev} & \multicolumn{1}{c}{\textbf{Eval}} \\

\midrule

\textcolor{gray!75}{0} & \multirow{6}{*}{1B} & \multicolumn{3}{c|}{\textcolor{gray!75}{Trained in a single stage using the loss defined in Eqn.~(\ref{eq:encsep})}}
& \textcolor{gray!75}{13.1} & \textcolor{gray!75}{11.8} & \textcolor{gray!75}{35.0} & \textcolor{gray!75}{33.9} & \textcolor{gray!75}{4.1} & \textcolor{gray!75}{4.0} & \textcolor{gray!75}{22.2} & \textcolor{gray!75}{23.7} \\
\cline{3-13} 

1 &  & 1st & SOT (Baseline) & $[\textbf{H}_p; \textbf{E}_t]$ & 12.4 & 11.3 & 39.8 & 39.1 & 4.6 & 4.6 & 21.5 & 21.6 \\


\textcolor{gray!75}{2} & & 2nd & \textcolor{gray!75}{SOT-CTC} & \textcolor{gray!75}{$[\textbf{H}_p; \textbf{E}_t]$} & \textcolor{gray!75}{14.8} & \textcolor{gray!75}{13.4} & \textcolor{gray!75}{39.1} & \textcolor{gray!75}{38.0} & \textcolor{gray!75}{5.5} & \textcolor{gray!75}{5.5} & \textcolor{gray!75}{23.9} & \textcolor{gray!75}{24.6}   \\

3 & & 3rd & SOP-based MT-ASR & $[\textbf{E}_{\text{sop}}; \textbf{H}_p; \textbf{E}_t]$ & \underline{11.8} & \underline{10.5} & \underline{29.6} & \underline{28.5} & \underline{3.9} & \underline{4.0} & 20.8 & 22.0 \\

\textcolor{gray!75}{4} & & 3rd & \quad \textcolor{gray!75}{- Mixed speech encoding} & \textcolor{gray!75}{$[\textbf{E}_{\text{sop}}; \textbf{E}_t]$} & \textcolor{gray!75}{33.1} & \textcolor{gray!75}{34.4} & \textcolor{gray!75}{76.6} & \textcolor{gray!75}{83.2} & \textcolor{gray!75}{16.3} & \textcolor{gray!75}{19.0} & \textcolor{gray!75}{74.9} & \textcolor{gray!75}{76.5} \\

\midrule
5 & \multirow{5}{*}{3B} & 1st & SOT (Baseline) & $[\textbf{H}_p; \textbf{E}_t]$ & 11.2 & 9.8 & 34.2 & 31.7 & 4.0 & 4.1 & 22.3 & 22.0 \\


\textcolor{gray!75}{6} & & 2nd & \textcolor{gray!75}{SOT-CTC} & \textcolor{gray!75}{$[\textbf{H}_p; \textbf{E}_t]$} & \textcolor{gray!75}{12.6} & \textcolor{gray!75}{11.1} & \textcolor{gray!75}{32.4} & \textcolor{gray!75}{30.7} & \textcolor{gray!75}{4.6} & \textcolor{gray!75}{4.7} & \textcolor{gray!75}{23.7} & \textcolor{gray!75}{23.4}  \\

7 & & 3rd & SOP-based MT-ASR & $[\textbf{E}_{\text{sop}}; \textbf{H}_p; \textbf{E}_t]$ & \underline{10.5} & \underline{9.2} & \underline{29.3} & \underline{28.1} & \underline{3.5}	& \underline{3.6} & \underline{17.0} & \underline{16.5} \\

\textcolor{gray!75}{8} & & 3rd & \quad \textcolor{gray!75}{- Mixed speech encoding} & \textcolor{gray!75}{$[\textbf{E}_{\text{sop}}; \textbf{E}_t]$} & \textcolor{gray!75}{70.9}	& \textcolor{gray!75}{85.1} & \textcolor{gray!75}{131.5} & \textcolor{gray!75}{154.0} & \textcolor{gray!75}{37.9} & \textcolor{gray!75}{50.3}  & \textcolor{gray!75}{109.7} & \textcolor{gray!75}{133.9}   \\

\midrule
9 & \multirow{5}{*}{8B} & 1st & SOT (Baseline, 1st-stage) & $[\textbf{H}_p; \textbf{E}_t]$ & 16.1 & 14.3 & 48.7 & 47.5 & 6.6 & 6.5 & 32.5 & 32.0 \\


\textcolor{gray!75}{10} & & 2nd & \textcolor{gray!75}{SOT-CTC} & \textcolor{gray!75}{$[\textbf{H}_p; \textbf{E}_t]$} & \textcolor{gray!75}{19.2} & \textcolor{gray!75}{17.4} & \textcolor{gray!75}{47.4} & \textcolor{gray!75}{45.5} & \textcolor{gray!75}{7.7} & \textcolor{gray!75}{7.6} & \textcolor{gray!75}{30.9} & \textcolor{gray!75}{30.3} \\

11 & & 3rd & SOP-based MT-ASR & $[\textbf{E}_{\text{sop}}; \textbf{H}_p; \textbf{E}_t]$ & \underline{15.0} & \underline{12.7} & \underline{40.0} & \underline{38.1} & \underline{5.3} & \underline{5.5} & \underline{24.4} & \underline{23.4} \\

\textcolor{gray!75}{12} & & 3rd & \quad \textcolor{gray!75}{- Mixed speech encoding} & \textcolor{gray!75}{$[\textbf{E}_{\text{sop}}; \textbf{E}_t]$} & \textcolor{gray!75}{53.6}	& \textcolor{gray!75}{57.2}	& \textcolor{gray!75}{120.7}	& \textcolor{gray!75}{138.9}	& \textcolor{gray!75}{30.2}	& \textcolor{gray!75}{34.0}	& \textcolor{gray!75}{84.5}	& \textcolor{gray!75}{102.1}    \\

\bottomrule[1.5pt]
\multicolumn{13}{r}{} \\ [-1.5ex]
\multicolumn{13}{r}{(\underline{Underline}: p-value $<$ 0.05 against corresponding baseline)} \\
\end{tabular}
\vspace{-10pt}
\label{table:proposed}
\end{table*}

\begin{table}[]
\vspace{-15pt}
\renewcommand{\arraystretch}{1.1}
\caption{Comparison between the proposed method and existing methods on the LibriMix datasets (270 hours for Libri2Mix and 186 hours for Libri3Mix, without any additional data augmentation). Word Error Rate (WER) is used for evaluation.}
\centering
\begin{tabular}{ll|cc|cc}
\toprule[1.5pt]
\multicolumn{2}{c|}{\multirow{2}{*}{\textbf{REF}}}
& \multicolumn{2}{c|}{\textbf{Libri2Mix}} & \multicolumn{2}{c}{\textbf{Libri3Mix}} \\
\cline{3-6}
& & \textbf{Dev} & \textbf{Eval} & \textbf{Dev} & \textbf{Eval} \\

\midrule

\multicolumn{1}{c|}{\multirow{8}{*}{\textbf{Noisy}}} & \multicolumn{5}{c}{\textbf{Without LLMs; with SSL for the speech encoder}} \\
\cline{2-6}

\multicolumn{1}{c|}{} & Training from Scratch \tablefootnote{\url{https://github.com/espnet/espnet/tree/master/egs2/librimix/sot_asr1} \label{myfoot}} & 19.4 & 17.1 & 30.5 & 28.2 \\

\multicolumn{1}{c|}{} & Conditional-Conformer \cite{guo21_interspeech} & 24.5 & 24.9 & - & - \\




\multicolumn{1}{c|}{} & TSE-V-Whisper \cite{10389752} & - & 12.0 & - & - \\

\multicolumn{1}{c|}{} & GEncSep \cite{shi_slt2024} & 17.2 & 15.0 & 28.0 & 25.9 \\

\cline{2-6}

\multicolumn{1}{c|}{} & \multicolumn{5}{c}{\textbf{With LLMs}} \\
\cline{2-6}

\multicolumn{1}{c|}{} & ID--5 & 11.2 & 9.8 & 34.2 & 31.7 \\

\multicolumn{1}{c|}{} & ID--7 & {10.5} & {9.2} & {29.3} & {28.1} \\

\midrule

\multicolumn{1}{c|}{\multirow{9}{*}{\textbf{Clean}}} & \multicolumn{5}{c}{\textbf{Without LLMs; with SSL for the speech encoder}} \\
\cline{2-6}

\multicolumn{1}{c|}{} & Training from Scratch \footnotemark[\value{footnote}] & 6.8 & 7.0 & 15.0 & 14.7 \\


\multicolumn{1}{c|}{} & W2V-Sidecar-ft. \cite{10095295} & 7.7 & 8.1 & - & - \\

\multicolumn{1}{c|}{} & WavLM-CLN \cite{10097139} & 7.1 & 7.6 & - & - \\

\multicolumn{1}{c|}{} & C-HuBERT LARGE \cite{10096630} & 6.6 & 7.8 & - & - \\


\multicolumn{1}{c|}{} & GEncSep \cite{shi_slt2024} & 6.4 & 6.6 & 13.3 & 13.1 \\

\cline{2-6}

\multicolumn{1}{c|}{} & \multicolumn{5}{c}{\textbf{With LLMs}} \\
\cline{2-6}

\multicolumn{1}{c|}{} & ID--5 & 4.0 & 4.1 & 22.3 & 22.0 \\

\multicolumn{1}{c|}{} & ID--7 & {3.5}	& {3.6} & {17.0} & {16.5} \\

\bottomrule[1.5pt]

\end{tabular}
\vspace{-5pt}
\label{table:method_comparision}
\end{table}

\subsection{Effect of the Proposed SOP-based MT-ASR}
Table~\ref{table:proposed} shows the performance of the proposed method on the LibriMix dataset. Both the noisy and clean sets were evaluated under two-talker and three-talker conditions. 
The first-stage training served as the baseline method, following the same setup as previous LLM-based MT-ASR approaches (except for the use of the task-definition prompt compared with \cite{meng2024large}). 
During the second-stage training, the SOT-CTC model experienced performance degradation. 
We argue that this degradation may be due to the presence of $<\text{blank}>$ tokens in CTC, which lead to sparse speech embeddings. 
In contrast, the baseline SOT system was trained with attention-based CE, which did not produce such sparse representations. 
In this work, we adopted multi-layer CNNs as down-sampling layers, and the sparsity introduced by CTC made it more difficult for the CNNs to extract meaningful information from the speech embeddings.

The negative impact caused by training the serialized CTC layers can be mitigated in the third training stage. 
The proposed SOP-based MT-ASR had significant improvements compared to both SOT and SOT-CTC. 
This indicates that SOP assists LLM decoding: explicitly providing guiding cues helps improve model performance. 
``$\text{– Mixed speech encoding}$'' experiments served as the ablation study to verify that the performance of MT-ASR was not due to the introduction of decoded information from the CTC outputs. 
``$\text{– Mixed speech encoding}$'' only fed the serialized CTC decoding results into the LLM, without the mixed speech encoding $\textbf{H}_p$. 
However, despite following the same training process (three-stage training, with only the SOP used as input in the third stage) as the SOP-based MT-ASR, it resulted in highly unstable performance, which is unacceptable. 
Besides, ID-0 represents the results of training the model in a single stage, which showed degraded performance across many evaluation sets. 
However, in the comparison across models with different parameter sizes, the 3B model achieved the best performance. 
Although the 8B model has more parameters, it did not lead to further improvement. 
We hypothesize that this is due to the limited amount of data, which makes it difficult for the model to learn effectively.

\begin{figure*}
    \centering
    \includegraphics[width=1.\linewidth]{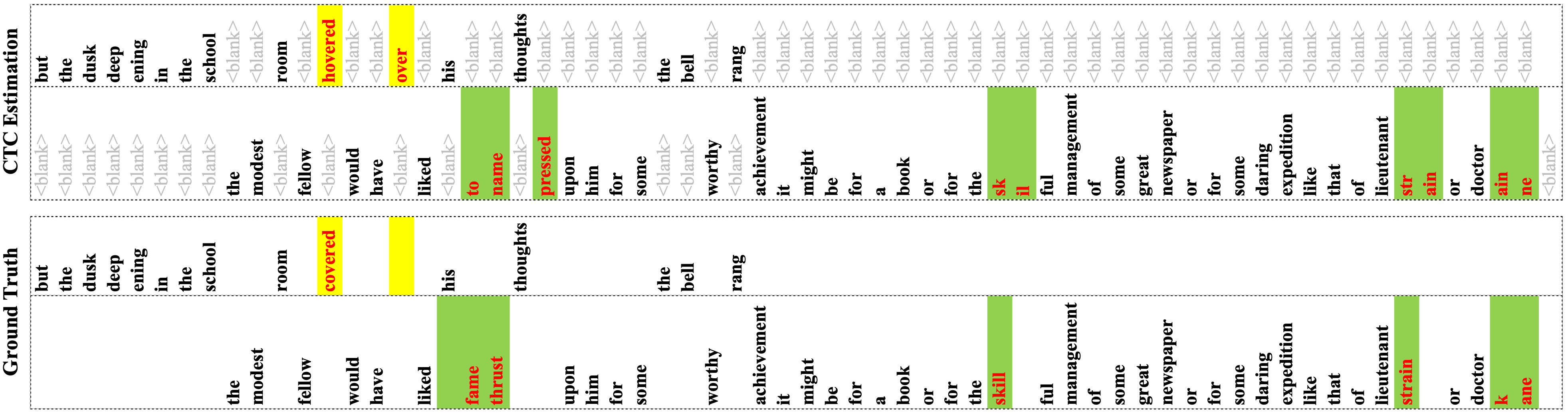}
    \vspace{-20pt}
    \caption{
        One example of the SOP content extracted using serialized CTC layers under the two-talker condition. The $<\text{blank}>$ frames were removed when all serialized CTC layers output blanks. 
        Positions marked in red indicate errors.
    }
    \label{fig:ctc-2spk}
\end{figure*}
\begin{figure*}
    \centering
    \includegraphics[width=1.\linewidth]{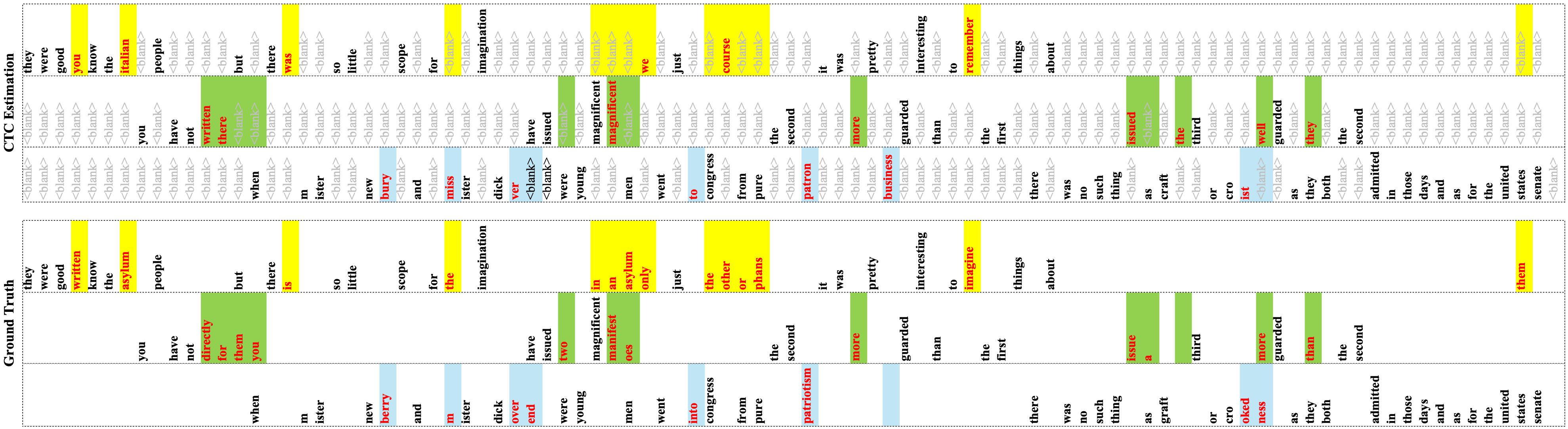}
    \vspace{-20pt}
    \caption{
        One example of the SOP content extracted using serialized CTC layers under the three-talker condition. The $<\text{blank}>$ frames were removed when all serialized CTC layers output blanks. 
        Positions marked in red indicate errors.  
    }
    \label{fig:ctc-3spk}
    \vspace{-10pt}
\end{figure*}

\subsection{Comparison Between Different Systems}
Table~\ref{table:method_comparision} shows the comparison between the proposed method and the existing methods on the LibriMix dataset. 
Existing LLM-based MT-ASR systems either augmented the LibriMix dataset \cite{shi2024advancing} or used synthesized training and testing sets instead of LibriMix \cite{meng2024large}. 
Compared with methods without LLMs, LLM-based approaches demonstrated significantly stronger performance on the development and evaluation sets of Libri2Mix (for both the noisy and clean sets). 
Even compared with ``TSE-V-Whisper'' \cite{10389752}, the SOT-LLM (ID-1 and ID-5) still performed better. 
This demonstrates that the strong contextual capabilities of LLMs are highly effective in handling the two-talker condition. 
However, LLM-based approaches underperformed on Libri3Mix relative to traditional AED E2E ASR systems. 
This may be because, in the 3Mix condition, the LLMs need to handle an excessive amount of information. 
SOT-MT-ASR systems without LLMs leverage cross-attention to fuse multi-modal information from speech embeddings and text embeddings. 
In contrast, LLM-based SOT-ASR systems rely solely on self-attention. 
Due to the presence of down-sampling layers and the projector layer, the speech embeddings are mapped into the representation space of text embeddings, resulting in the loss of some speech-specific information. 
Compared to the two-talker scenario, the three-talker condition is more complex, and relying only on text-embedding information may make it difficult to properly align the speech content of different speakers. 

\subsection{Effect of SOP for Speech Encoding}
We analyzed how the serialized CTC layers performed. 
Fig.~\ref{fig:ctc-2spk} and Fig.~\ref{fig:ctc-3spk} show the examples of serialized CTC layers under two-talker and three-talker conditions, respectively. 
A clear experimental observation was that regions with high overlap, where multiple talkers produced CTC outputs at the same time step, or where frequent switching occurred between different talkers’ CTC outputs across adjacent frames tended to result in more prediction errors. 
Besides, the overall output quality of the serialized CTC provides complete and well-aligned speech content for different talkers.

\section{Conclusions}
\label{sec:conclu}
In this paper, we proposed a serialized output prompting (SOP) method to explicitly guide LLMs for multi-talker (MT) ASR. 
To extract serialized MT content from mixed speech representations, we inserted a Separator and speaking-time-aligned serialized CTC layers after the speech encoder. 
Each CTC branch corresponds to an individual talker, with outputs ordered according to their speaking starting time. 
The SOP was then generated by applying greedy decoding to the serialized CTC outputs and is subsequently used as an explicit prompt to guide the LLM decoder. 
To enable effective learning, we introduce a three-stage training strategy comprising:
(1) fine-tuning with SOT, 
(2) extraction of talker-specific serialized information, and
(3) SOP-based adaptation. 
Experimental results on the LibriMix dataset indicate that the speech encoder implicitly performed a degree of temporal re-alignment, even prior to explicit separation. 
Despite this, the serialized CTC layers produced good-quality outputs. 
Furthermore, the proposed SOP-based MT-ASR system demonstrated substantial performance gains over the baseline SOT model, highlighting the effectiveness of using serialized prompts to guide LLM-based decoding in MT scenarios.

\clearpage
\bibliographystyle{IEEEtran}
\bibliography{ref}

\end{document}